\documentclass[a4paper, conference]{IEEEtran}

\usepackage{cite}

\ifCLASSINFOpdf
  \usepackage{graphicx}
  \graphicspath{{./images/}}
\else
  \usepackage{graphicx}
  \graphicspath{{./images/}}
\fi
\usepackage{amsmath}
\usepackage{amssymb}

\ifCLASSOPTIONcompsoc
\else
 \usepackage[caption=false,font=footnotesize]{subfig}
\fi

\usepackage{color}

\usepackage{amsmath}
\begin{document}

\title{Predicting Online Video Advertising Effects \\ with Multimodal Deep Learning}

\author{
  \IEEEauthorblockN{
    Jun Ikeda\IEEEauthorrefmark{1},
    Hiroyuki Seshime\IEEEauthorrefmark{2},
    Xueting Wang\IEEEauthorrefmark{1},
    and
    Toshihiko Yamasaki\IEEEauthorrefmark{1}
  }
  \IEEEauthorblockA{
    \IEEEauthorrefmark{1}The University of Tokyo, 7--3--1 Hongo, Bunkyo-ku, Tokyo, 113--8656 Japan \\
    \IEEEauthorrefmark{1}\{ikeda, xt\_wang, yamasaki\}@hal.t.u-tokyo.ac.jp
  }
  \IEEEauthorblockA{
    \IEEEauthorrefmark{2}Septeni Co., Ltd. 8--17--1 Nishishinjuku, Shinjuku-ku, Tokyo, 160--6128 Japan \\
    \IEEEauthorrefmark{2}h\_seshime@septeni.co.jp
  }

    % \IEEEauthorblockN{Jun Ikeda and Xueting Wang and Toshihiko Yamasaki and Kiyoharu Aizawa}
    % \IEEEauthorblockA{The University of Tokyo, Hongo 7--3--1, Bunkyo-ku, Tokyo, Japan}
}

\maketitle

\begin{abstract}
    With expansion of the video advertising market, research to predict the effects of video advertising is getting more attention.
    Although effect prediction of image advertising has been explored a lot, prediction for video advertising is still challenging with seldom research.
    In this research, we propose a method for predicting the click through rate (CTR) of video advertisements and analyzing the factors that determine the CTR.
    In this paper, we demonstrate an optimized framework for accurately predicting the effects by taking advantage of the multimodal nature of online video advertisements including video, text, and metadata features.
    In particular, the two types of metadata, i.e., categorical and continuous, are properly separated and normalized.
    To avoid overfitting, which is crucial in our task because the training data are not very rich, additional regularization layers are inserted.
    Experimental results show that our approach can achieve a correlation coefficient as high as 0.695, which is a significant improvement from the baseline (0.487).
\end{abstract}

% keywords

\IEEEpeerreviewmaketitle

\section{Introduction}

With the development of high-speed, large-capacity, and low-latency wireless communication environments, online video viewership on smartphones has been increasing.
Video advertisements embedded with such content are becoming popular among users.
Hence, video ads have become an established advertising method among companies, and video ads posted on  social networking service  apps such as Facebook and Instagram, which have a large user base, are increasing.
According to the survey\footnote{CyberAgent conducted market research on domestic video advertising in 2019. https://www.cyberagent.co.jp/news/detail/id=24125, Dec. 2019.} from the Cyber Agent Inc., the size of the video advertising market in 2019 was forecasted to be 141\% compared to that in the preceding year; in particular, the demand for smartphone video advertising grew to 147\% compared to the year before that, and accounted for approximately 89\% of the total video advertising market.

With expansion of the video advertising market, research to predict the effects of video advertising is needed, because ad creators want to know the effects of their ads and how they can create more effective ads.
Research on image advertising has already been actively conducted.
In these studies, many researchers used the click through rate (CTR), which is the ratio of the frequency users click on a video to the number of users to whom the video is shown (\ref{eq:ctr}), as an indicator of the effects of advertising on users.
\begin{equation}
    \label{eq:ctr}
    {\rm CTR} = \frac{\rm Number ~ of ~ clicks}{\rm Number ~ of ~ impressions}.
\end{equation}

Although many studies have worked on the effects of image advertising~\cite{iwazaki2018,xia2019ctr,xia2020}, there are seldom research on video advertising.
In particular, to our knowledge, research on predicting the CTR for online video ads has not been conducted extensively.
CTR prediction enables the recommendation of videos with higher advertising effects and support for producing such video ads.

Hence, the goal of this research is to predict the impact of online video advertising from the data available before release and to identify the factors that contribute to the determination of the CTR, and thus, finally to provide the basic tools that support the production of video advertising.
We design the manner of feature extraction of metadata and the suppression of overfitting so as to take advantage of the multimodal nature of online video ads.
As a result, we enhance the accuracy of prediction considerably from the baseline.

\section{Related works}

\subsection{CTR prediction}

Studies on predicting the CTR of image advertising have been actively conducted and have attracted much attention in the advertising industry.
Early studies on CTR prediction used logistic regression as a prediction model to predict CTR from metadata such as page URLs and keywords\cite{richardson2007predicting,chakrabarti2008contextual}.
Zhang et al.\cite{zhang2016deep} introduced deep learning into CTR prediction and further used factorization machine (FM)~\cite{rendle2010factorization} to account for the interrelationship between metadata.
In recent years, DeepFM~\cite{huifeng2017deepfm} and xDeepFM~\cite{lian2018xdeepfm} have been developed as extensions of the FM. These methods dealt primarily with metadata.
Chen et al.\cite{chen2016deep} showed the effectiveness of using features extracted from images by convolutional neural networks (CNNs) for prediction; Iwazaki\cite{iwazaki2018} proposed a method to use features from text extracted by CNNs in addition to images and metadata; Xia et al.\cite{xia2019ctr, xia2020} showed that extracting features from embedded text improved prediction accuracy.
In addition, they visualized the contribution of images, metadata, and text features to the prediction by introducing an attention mechanism\cite{AttentionisAllYouNeed}.

% Richardson et al.\cite{richardson2007predicting} and Chakrabarti et al.\cite{chakrabarti2008contextual} estimated the CTR by logistic regression using information such as landing page URLs, search keywords, and titles as inputs.
% Rendle et al.\cite{rendle2010factorization} significantly improved the learning speed and improved the accuracy of CTR prediction using an algorithm called FTRFL that generates a sparse model.
% Zhang et al. \cite{zhang2016deep} introduced the FTRFL proposed by Rendle et al., succeeded in making the deep neural network (DNN) function efficiently, and proposed a better model than before.
% Zhou et al.\cite{zhou2018deep} also showed that using different expression vectors for each ad based on user's interests significantly improved the expressiveness of the model and improved the CTR prediction accuracy.
% Chen et al.\cite{chen2016deep} entered multimodal information such as advertising images and metadata into the network to improve the accuracy of CTR prediction.
% Iwazaki et al.\cite{iwazaki2018} have shown that entering text, in addition to advertising images and metadata, improves the accuracy of CTR prediction.
% Xia et al.\cite{xia2019ctr} improved the accuracy of CTR prediction by cutting out and inputting advertising images to multiple fixed sizes, allowing the model to consider the features of multiple locations in the image. Furthermore, by introducing an attention mechanism\cite{AttentionisAllYouNeed}, they visualized the contribution of images, metadata, and text features to the prediction.

\subsection{Prediction of TV commercial impressions}

\begin{figure}[tb]
  \centering
  \includegraphics[width=.9\linewidth]{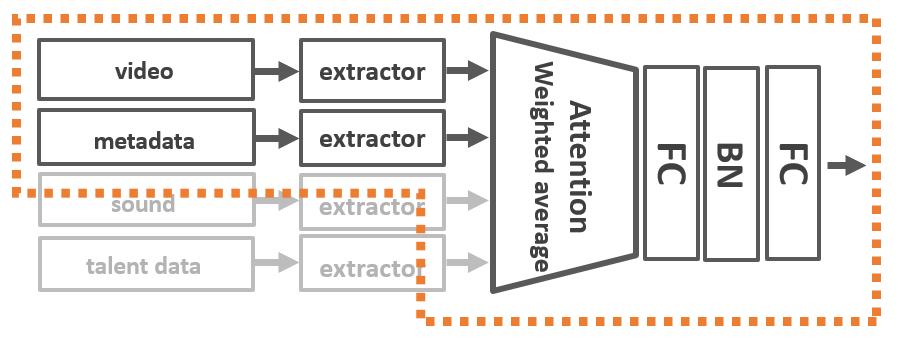}
  \caption{Overview of the model of Nakamura et al.\cite{nakamura2018cnn}. The baseline model in this research is the modified model, which partially includes the model of Nakamura et al.\cite{nakamura2018cnn}, surrounded by the broken orange line.}
  \label{fig:nakamura_model}
\end{figure}

Nakamura et al.\cite{nakamura2018cnn} estimated the impressions of video advertisements and analyzed the factors affecting the CTR for 15s TV commercials.
They predicted four impressional and emotional effects. These are quantified as values from 0 to 1 based on the questionnaire.
% They predicted the following four impressional and emotional effects:
% \begin{itemize}
%   \item Recognition rate: how much participants remember the advertisement.
%   \item Willingness rate: how much participants feel like buying the product/service.
%   \item Interesting rate: how much participants become interested in the product/service.
%   \item Favorableness rate: how much participants like the content of the advertisement itself.
% \end{itemize}
The impressions were estimated using the video and audio of TV commercials, metadata such as product classifications and broadcast patterns, and information on the talents who appeared in the commercials.
The multimodal data were integrated by the neural network (Fig.~\ref{fig:nakamura_model}), and attractiveness of the commercials were estimated.
Further, comparison of the magnitude of the weightage  calculated by the attention mechanism, the amount of video and audio content, metadata, and talent information contributed to the estimation of attractiveness.
Relatively high correlation coefficients of 0.73, 0.67, 0.80, and 0.63 between the predicted values and ground truth for each recognition, favor, purchase arousal, and interest were obtained, demonstrating that high accuracy can be obtained by combining multimodal data.
% Li et al. \cite{li2019dnn} improved the prediction accuracy based on this research by adding text information including text and narration of video advertisements.

Although this study succeeded in predicting the effects of video ads with relatively high accuracy and analyzing the factors, our preliminary experiment demonstrated that the targeted CTR prediction accuracy obtained using this model for online video ads was not so high -- the correlation coefficient therein was 0.487.
This might be due to the differences in the characteristics of TV commercials (target of~\cite{nakamura2018cnn}) and online video ads (target of this research).
For example, online video ads have multiple metadata with different scales. Moreover, there are many similar ads with different metadata in some sections.
Therefore, in order to obtain high prediction accuracy, it is necessary to optimize the architecture and hyperparameters for online video ads.

\section{Proposed Methods}
\begin{figure}
  \centering
  \includegraphics[width=.9\linewidth]{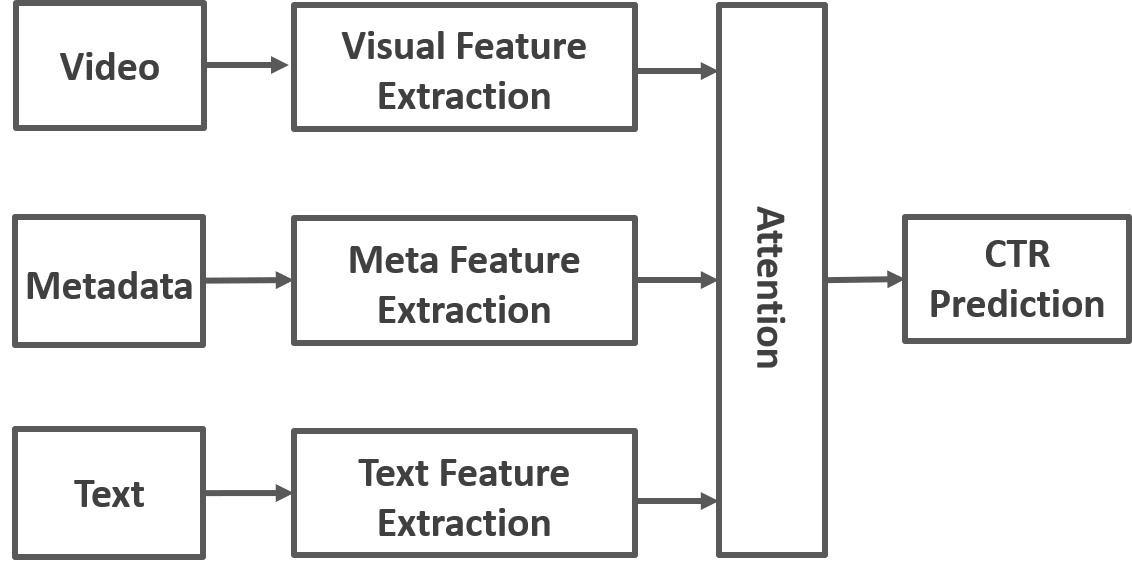}
  \caption{Overview of proposed network.}
  \label{fig:overview}
\end{figure}

Fig.~\ref{fig:overview} shows the overall diagram of the proposed regression model using the CNN for CTR prediction.
The features of video frames, metadata, and text data are individually extracted, and each feature is passed through the full connection (FC) layer to fix the feature dimension.
Subsequently, the feature vectors are weighted by the attention mechanism, and the weighted average of the multimodal feature vectors is calculated. Based on this feature vector, the CTR is predicted by the neural network.
The following describes each feature extraction network and prediction network from the integrated modal feature.

\subsection{Visual feature extraction}

\begin{figure}[!t]
  \centering
  \includegraphics[width=.9\linewidth]{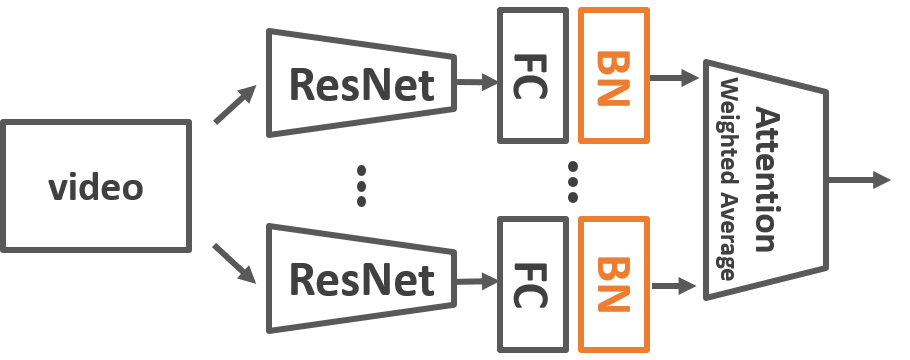}
  \caption{Network for visual feature extraction.}
  \label{fig:frame_feature_extraction_network}
\end{figure}

From each video, $n$ frames were extracted so that the interval between each frame was the same.
Each frame was resized to $224 \times 224$ and input into ResNet50~\cite{Kaiming2015ResNet}, which has achieved very high accuracy in image classification tasks, pre-trained on ImageNet~\cite{Deng2009imagenet}, in order to extract the features of each frame.
The last layer of ResNet50 was removed to obtain the features of the hidden layer, and all parameters were fixed. The size of the extracted features was reduced from 2048 to 256 by the Fully Connected (FC) layer.
Subsequently, the features were normalized by the batch normalization (BN)~\cite{Sergey2015BatchNorm} layer.
This is one of the modifications for suppressing overfitting. Finally, $n$ frame feature vectors $ f^i_{\rm frame}\{i=1,...,n\} $ were obtained.
In order to integrate these feature vectors, the attention mechanism calculated weights $ \alpha^i\{i=1,...,n\} $ for the input vectors, and according to these weights, the weighted average $F_{\rm visual}$ was calculated (\ref{eq:f_attention1}) as an integrated feature of 15 frames.
This attention mechanism enhances the prediction accuracy because it allows the model to selectively use the $n$ frame features.
Additionally, visualization of the attention weights made it possible to analyze which frames are more effective in predicting the CTR.
\begin{align}
  \label{eq:f_attention1}
  & F_{\rm visual} = \sum_{i=1}^{n} \alpha^i f^i_{\rm frame}, \\
  \label{eq:f_attention2}
  & \alpha^i = \psi_\alpha(f^1_{\rm frame}, f^2_{\rm frame}, ..., f^{n}_{\rm frame}).
\end{align}
Here, $ \psi_\alpha $ represents the attention mechanism receiving frame features $f^i_{\rm frame}\{i=1, ..., n\}$ as the input and outputting the weights of each frame $\alpha^i$.

\subsection{Metadata feature extraction}
There are two kinds of metadata: qualitative/categorical variables and quantitative/continuous variables.
In the existing method, qualitative variables are one-hot encoded and concatenated with each other, and then concatenated with quantitative variables.
Subsequently, they are inputted into the feature extraction network.
In our model, we input qualitative variables and quantitative variables separately.
The qualitative variables, which were one-hot encoded and concatenated, were input to the FC and BN layers to acquire 16-dimensional feature vectors.
At the same time, the quantitative variables were input to the FC and BN layers, and 240-dimensional feature vectors were obtained.
Then, both vectors were normalized respectively by a BN layer.
Both vectors were concatenated and input to the next layers.
We finally obtained a 256-dimensional feature vector.

\begin{figure}[!t]
  \centering
  \includegraphics[width=.9\linewidth]{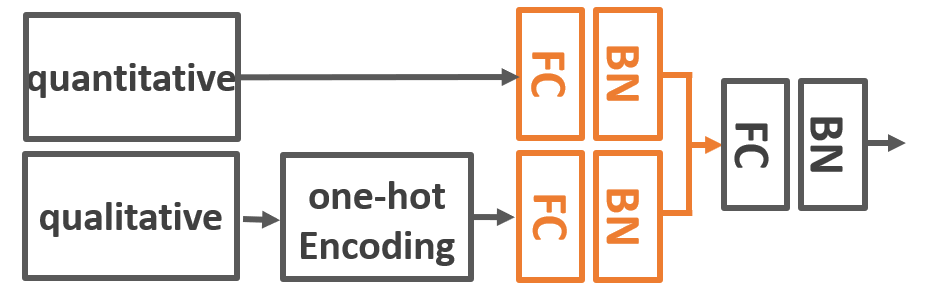}
  \caption{Network for metadata feature extraction.}
  \label{fig:meta_network}
\end{figure}

\subsection{Text feature extraction}
The text data were embedded into 300-dimensional vectors by Doc2Vec\cite{lau2016empirical}.
Before training the doc2vec model, we preprocess texts with MeCab, which is a Japanese morphological analysis engine, to include a space between two words\cite{xia2019ctr}.
We use the doc2vec model implemented in Gensim\cite{gensim2010}.
The sum of the embedded vectors was input into the next layers, and 256-dimensional feature vectors were obtained.
The batch normalization layers were inserted to suppress overfitting.

\begin{figure}[!t]
  \centering
  \includegraphics[width=.9\linewidth]{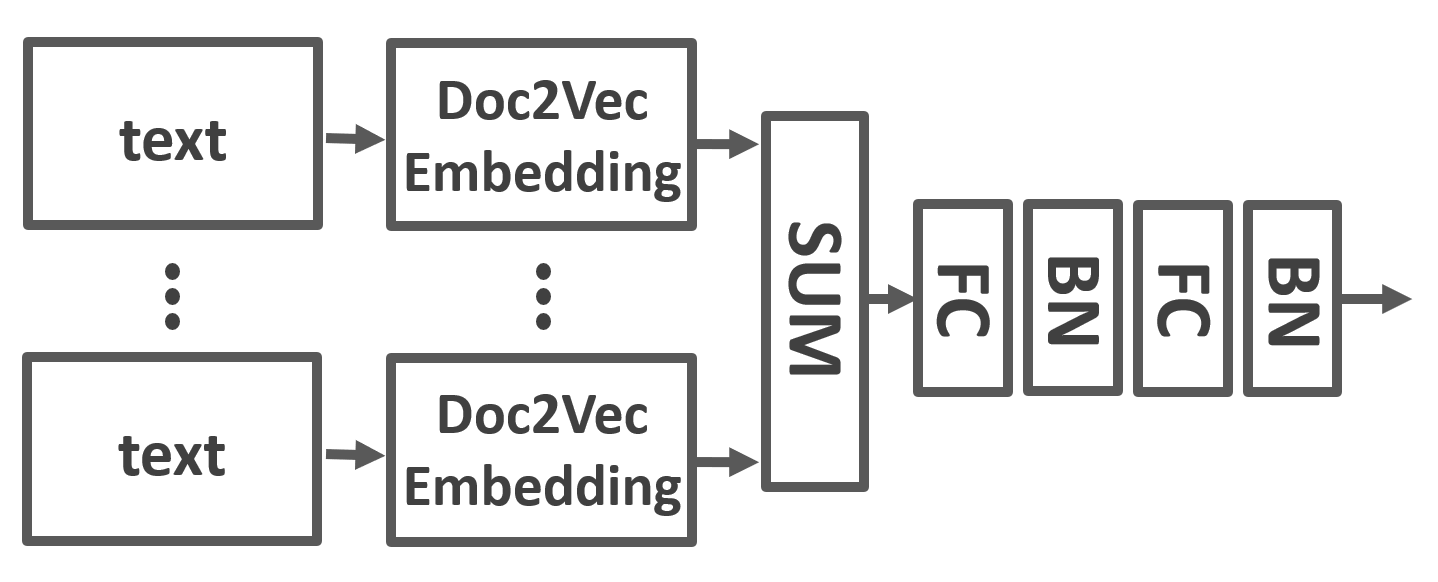}
  \caption{Network for text feature extraction.}
  \label{fig:text_feature_extraction}
\end{figure}

\subsection{Integration of multimodal features}

\begin{figure}[!t]
  \centering
  \includegraphics[width=.9\linewidth]{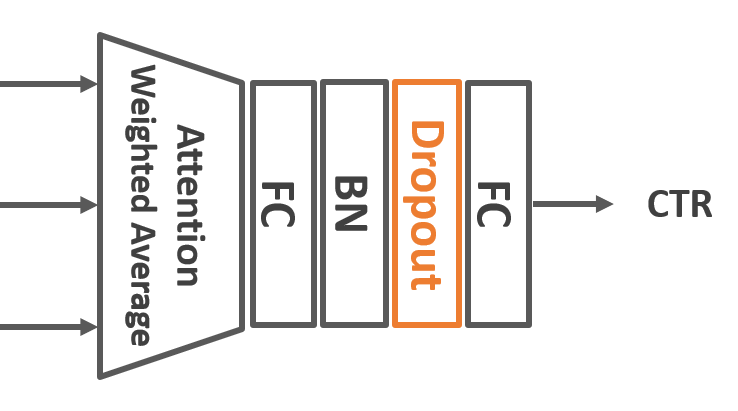}
  \caption{Network for integration of features and CTR prediction.}
\end{figure}

The weighted average $ F $ of the feature vectors that were normalized by their own size was calculated according to the weight calculated by the attention mechanism as an ad feature vector.
\begin{align}
  \label{eq:ad_attention1}
  & F = \beta_{\rm visual} \frac{F_{\rm visual}}{\|F_{\rm visual}\|} + \beta_{\rm meta} \frac{F_{\rm meta}}{\|F_{\rm meta}\|} + \beta_{\rm text} \frac{F_{\rm text}}{\|F_{\rm text}\|}, \\
  \label{eq:ad_attention2}
  & \beta_{\rm visual}, \beta_{\rm meta}, \beta_{\rm text} = \psi(\frac{F_{\rm visual}}{\|F_{\rm visual}\|}, \frac{F_{\rm meta}}{\|F_{\rm meta}\|}, \frac{F_{\rm text}}{\|F_{\rm text}\|}),
\end{align}
where $\psi_\beta$ represents the attention mechanism receiving modal features $F_{\rm visual}, F_{\rm meta}, F_{\rm text}$ as the input and outputting the weights of each modal feature $\beta_{\rm visual}, \beta_{\rm meta}, \beta_{\rm text}$.

\subsection{CTR prediction}
The ad feature vector was input into the FC, BN, and dropout (probability 0.5)~\cite{dropout} layers, and finally input into the FC layer to obtain the one-dimensional CTR.
A dropout layer was then newly added to suppress overfitting.

\section{Experiments}

\begin{table}[tb]
  \centering
  \caption{Number of data}
  \label{tab:n_data}
  \begin{tabular}{cccc}
    \hline
    & train & valid & test \\ \hline
    Number of data & 80,771 & 8,061 & 9,655 \\
    Number of unique videos & 20,618 & 2,032 & 2,945 \\
    \hline
  \end{tabular}
\end{table}

\begin{table}[!t]
  \centering
  \caption{List of Metadata}
  \label{tab:metadata}
  \begin{tabular}{lcccc}
    \hline
    Key  & Qualitative & Quantitative & Text & Input \\ \hline
    promotion\_id &  \checkmark &  &  & \checkmark \\
    publisher\_platform &  \checkmark &  &  & \checkmark \\
    platform &  \checkmark &  &  & \checkmark \\
    genre & \checkmark &  &  & \checkmark \\
    sub\_genre & \checkmark &  &  & \checkmark \\
    web\_app & \checkmark &  &  & \checkmark \\
    funnel & \checkmark &  &  & \checkmark \\
    creative\_type & \checkmark &  &  & \checkmark \\
    targeting\_type & \checkmark &  &  & \checkmark \\
    targeting\_gender & \checkmark &  &  & \checkmark \\
    targeting\_os & \checkmark &  &  & \checkmark \\
    targeting\_device & \checkmark &  &  & \checkmark \\
    targeting\_age\_min & & \checkmark &  & \checkmark \\
    targeting\_age\_max & & \checkmark &  & \checkmark \\
    target\_cost & & \checkmark &  & \checkmark \\
    target\_cpa & & \checkmark &  & \checkmark \\
    advertiser\_name & &  & \checkmark & \checkmark \\
    account\_name & &  & \checkmark & \checkmark \\
    promotion\_name & &  & \checkmark & \checkmark \\
    creative\_title & &  & \checkmark & \checkmark \\
    creative\_description & &  & \checkmark & \checkmark \\
    impressions &  & \checkmark &  &  \\
    clicks & & \checkmark &  & \\ \hline
  \end{tabular}
\end{table}

\begin{table*}
  \caption{Results comparison of the proposed methods and baseline methods. \label{tab:results}}
  \centering
  \begin{tabular}{clcccccccc}
    \hline
     & & \multicolumn{3}{c}{Input} & \multicolumn{2}{c}{Metadata} & Additional Layer & \multicolumn{2}{c}{Metrics} \\
    ID & Method & Visual & Metadata & Text & Normalized & Separated & Regularization & RMSE & R \\ \hline
    1 & Visual & \checkmark &  &  &  & \checkmark & \checkmark & 0.132 & 0.485 \\
    2 & Metadata &  & \checkmark &  &  & \checkmark & \checkmark & 0.119 & 0.618 \\
    3 & Text &  &  & \checkmark &  & \checkmark & \checkmark & 0.149 & 0.334 \\ \hline
    4 & Visual \& Metadata & \checkmark & \checkmark &  &  & \checkmark & \checkmark & 0.109 & 0.684 \\
    5 & Visual \& Text & \checkmark &  & \checkmark &  & \checkmark & \checkmark & 0.127 & 0.540 \\
    6 & Metadata \& Text &  & \checkmark & \checkmark &  & \checkmark & \checkmark & 0.115 & 0.637 \\ \hline
    7 & Unprocessed metadata & \checkmark & \checkmark & \checkmark &  &  & \checkmark & 0.126 & 0.540 \\
    8 & Normalized metadata & \checkmark & \checkmark & \checkmark & \checkmark &  & \checkmark & 0.113 & 0.656 \\
    9 & Separated \& normalized metadata & \checkmark & \checkmark & \checkmark & \checkmark & \checkmark & \checkmark & 0.110 & 0.675 \\ \hline
    10 & No Additional Layer for suppressing overfitting & \checkmark & \checkmark & \checkmark &  & \checkmark &  & 0.121 & 0.598 \\ \hline
    11 & Ours ({\it n} = 10) & \checkmark & \checkmark & \checkmark &  & \checkmark & \checkmark & 0.111 & 0.671 \\
    12 & Ours ({\it n} = 20) & \checkmark & \checkmark & \checkmark &  & \checkmark & \checkmark & 0.111 & 0.667 \\ \hline
    13 & baseline \cite{nakamura2018cnn} & \checkmark & \checkmark &  &  &  &  & 0.131 & 0.487 \\
    14 & \textbf{Ours (without Text)} & \checkmark & \checkmark &  &  & \checkmark & \checkmark & \textbf{0.109} & \textbf{0.684} \\
    15 & \textbf{Ours} & \checkmark & \checkmark & \checkmark &  & \checkmark & \checkmark & \textbf{0.107} & \textbf{0.695} \\ \hline
  \end{tabular}
\end{table*}

\subsection{Dataset}

In this research, we used the online video ad data for CTR prediction that were actually used in a business by Septeni Co., Ltd.
These data contain video ads distributed on Facebook and Instagram since January 2018 until December 2019.

The video ad data that satisfy the three conditions described later are split into training, validation, and test datasets.
The ads in the validation and test datasets should be newer than those in the training dataset to ensure the accuracy of CTR prediction for future ads, because the goal of this research is to predict the CTR of future ads.
Therefore, the data were split in chronological order.
In addition, the dataset includes many overlapping data that have the same video content and slightly different metadata.
Therefore, the data were split so that ads with the same video content were not separated into the training and test datasets, because it is easier for the model to predict the CTR of data similar to training data and the prediction accuracy is higher than expected.
The number of data and unique videos in the dataset were counted and are shown in Table~\ref{tab:n_data}.

Only data that satisfied the following three conditions were used as the dataset in this research.
\begin{itemize}
  \item Displayed more than 500 times
  \item One or more clicks
  \item 5-30s videos
\end{itemize}

The CTR of ads displayed only a few times is unstable, because when the number of impressions (denominator of the CTR) is small, a small change in the number of clicks (the numerator) greatly changes the CTR.
Therefore, only ads displayed more than 500 times are used in the dataset.
Furthermore, only ads that were clicked on once or more times were used in the dataset, because the CTR of never-clicked data is always 0 regardless of the number of impressions, and the concentration of the CTR distribution hinders the network training.

\begin{figure}[!t]
  \centering
  \includegraphics[width=.9\linewidth]{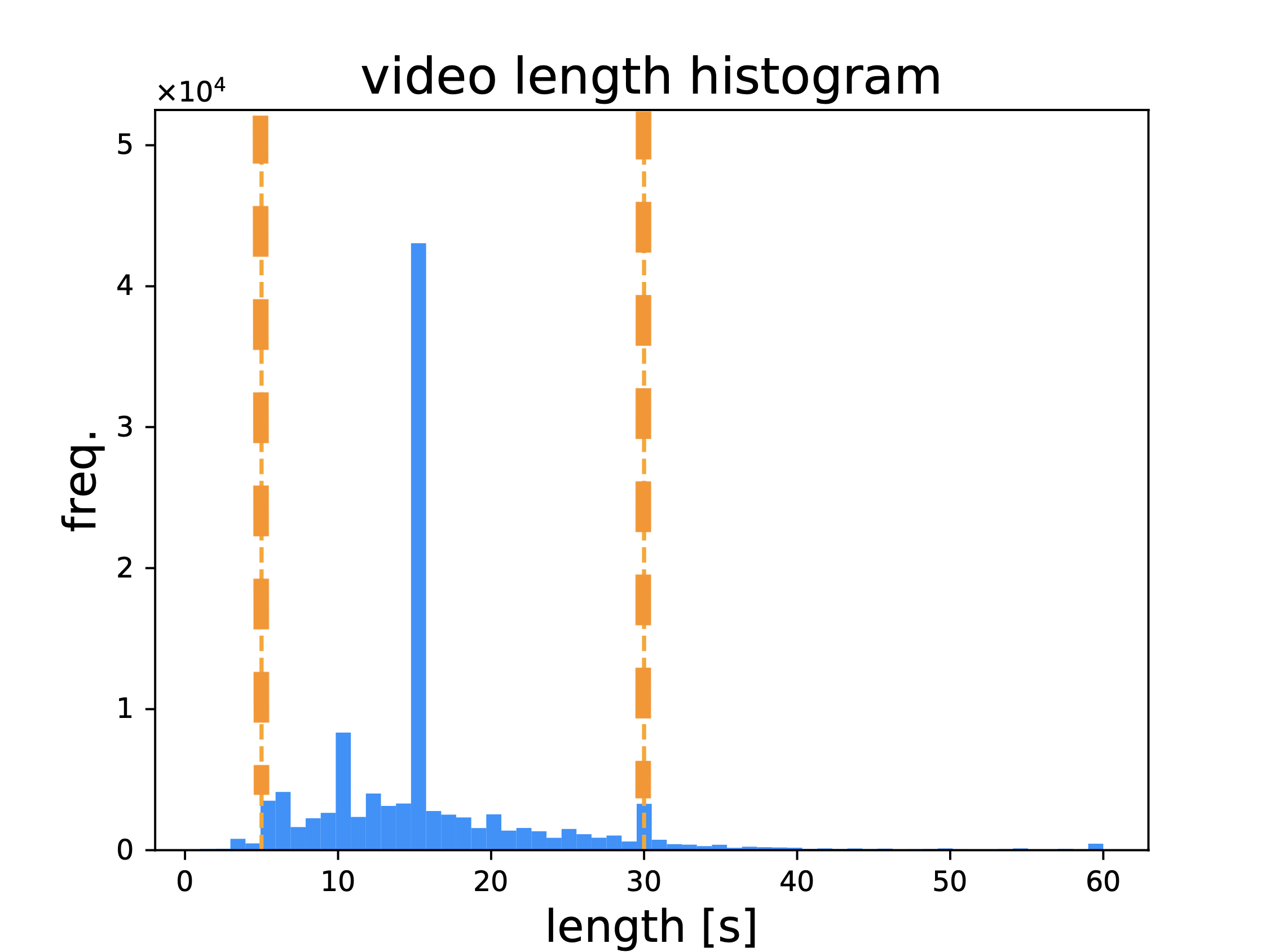}
  \caption{Lengths of videos. The two dotted lines indicate 5 and 30.}
  \label{fig:video_length}
\end{figure}

The lengths of videos included in the dataset ranged from as short as a second to as long as minutes.
However, as shown in Fig.~\ref{fig:video_length}, most videos were 5 to 30s long, and the video lengths were concentrated at 15s in this dataset; consequently, only 5-30s videos were used in the dataset.

\subsection{Metadata}
The video ad data contain many kinds of metadata.
These metadata are classified into qualitative variables, quantitative variables, and text data.
The kinds of metadata and the classification are listed in Table~\ref{tab:metadata}.
The aim of this work is to predict the CTR before releasing the video ads.
Therefore, metadata that are available before the release are only used.

\subsection{Network training}
The loss function is the mean squared error, and the optimizer is the momentum stochastic gradient descent.
The momentum is set to 0.9. The learning epoch is 200.
In every epoch, prediction is performed on the validation dataset in order to adopt the model that scores the minimum MSE among the 200 epochs as the best model.
The logarithmically transformed CTR in (\ref{eq:ctr_translation}) was used for learning.
\begin{equation}
  CTR = \log_{10} ({100 ~ CTR_{\rm raw} + 1}).
  \label{eq:ctr_translation}
\end{equation}

\begin{figure*}[!t]
  \centering
  \begin{tabular}{c}
    % frame
    \begin{minipage}{0.45\hsize}
      \centering
      \includegraphics[scale=0.39]{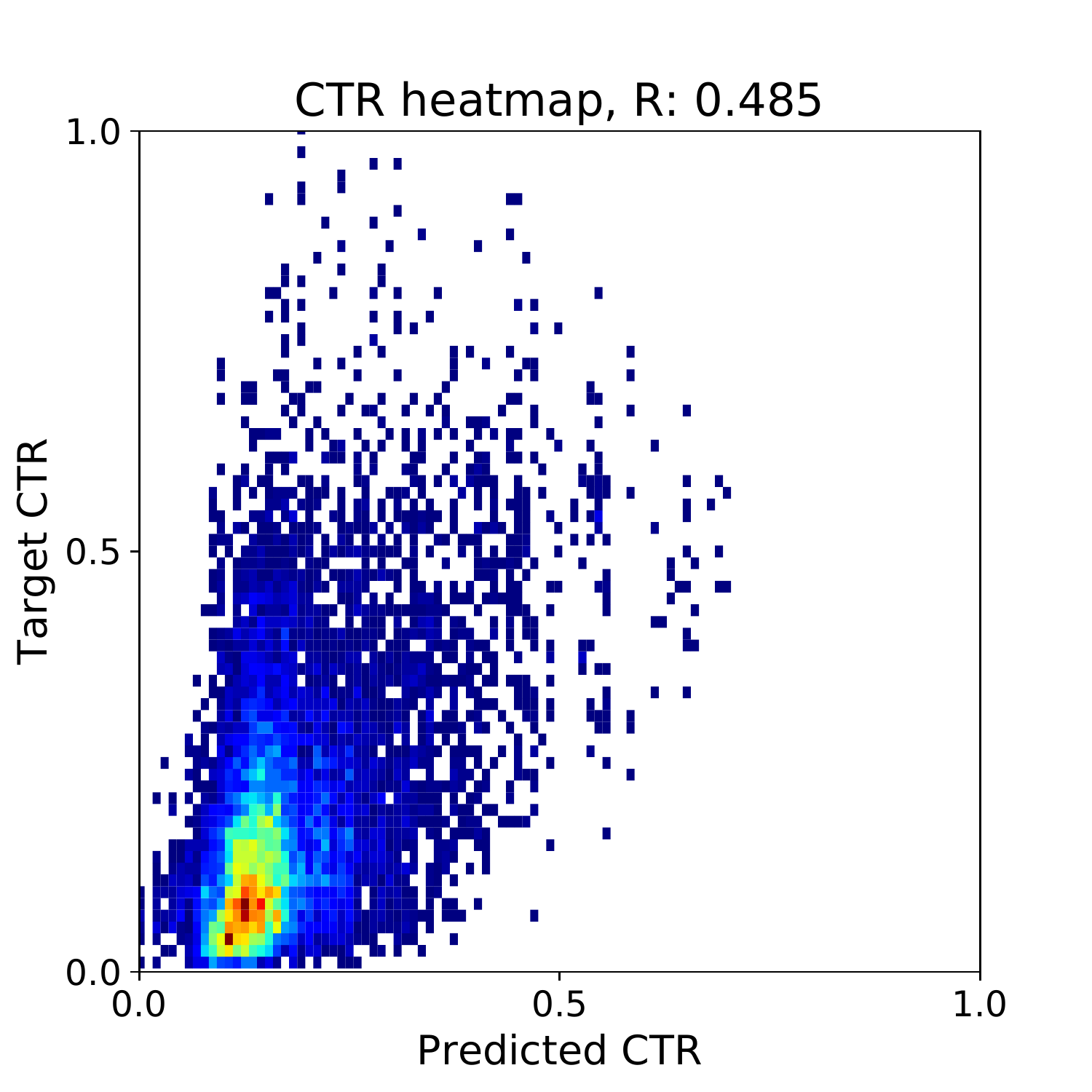}
      \begin{center}
        (a) Visual
      \end{center}
      % \caption{Scatter of the ground truth and predicted valus of single feature model (Visual).}
      % \subcaption{Frame}
      \label{fig:frame}
    \end{minipage}

    % meta
    \begin{minipage}{0.45\hsize}
      \centering
      \includegraphics[scale=0.39]{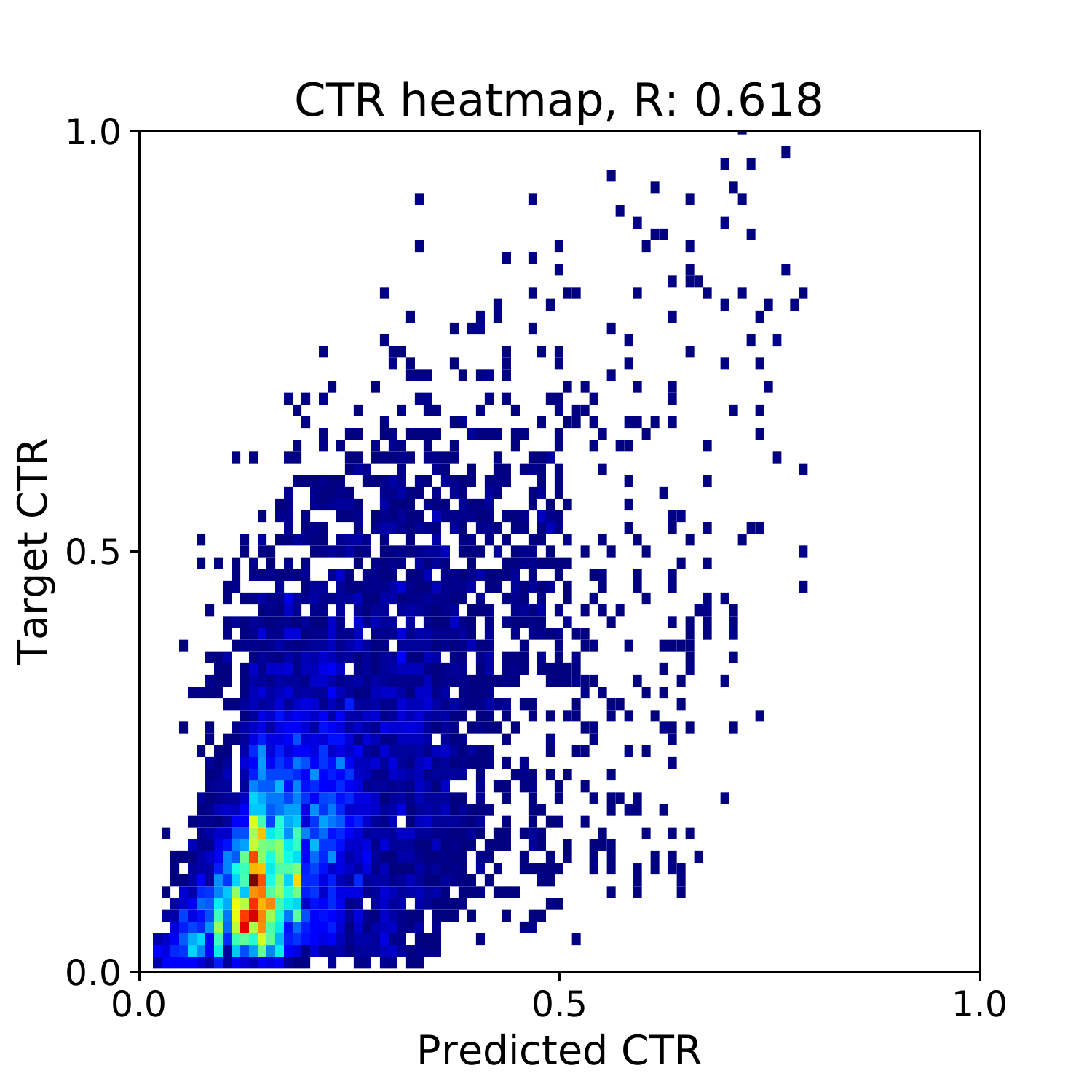}
      \begin{center}
        (b) Metadata
      \end{center}
      % \subcaption{Meta}
      \label{fig:meta}
    \end{minipage}
    \\
    % text
    \begin{minipage}{0.45\hsize}
      \centering
      \includegraphics[scale=0.39]{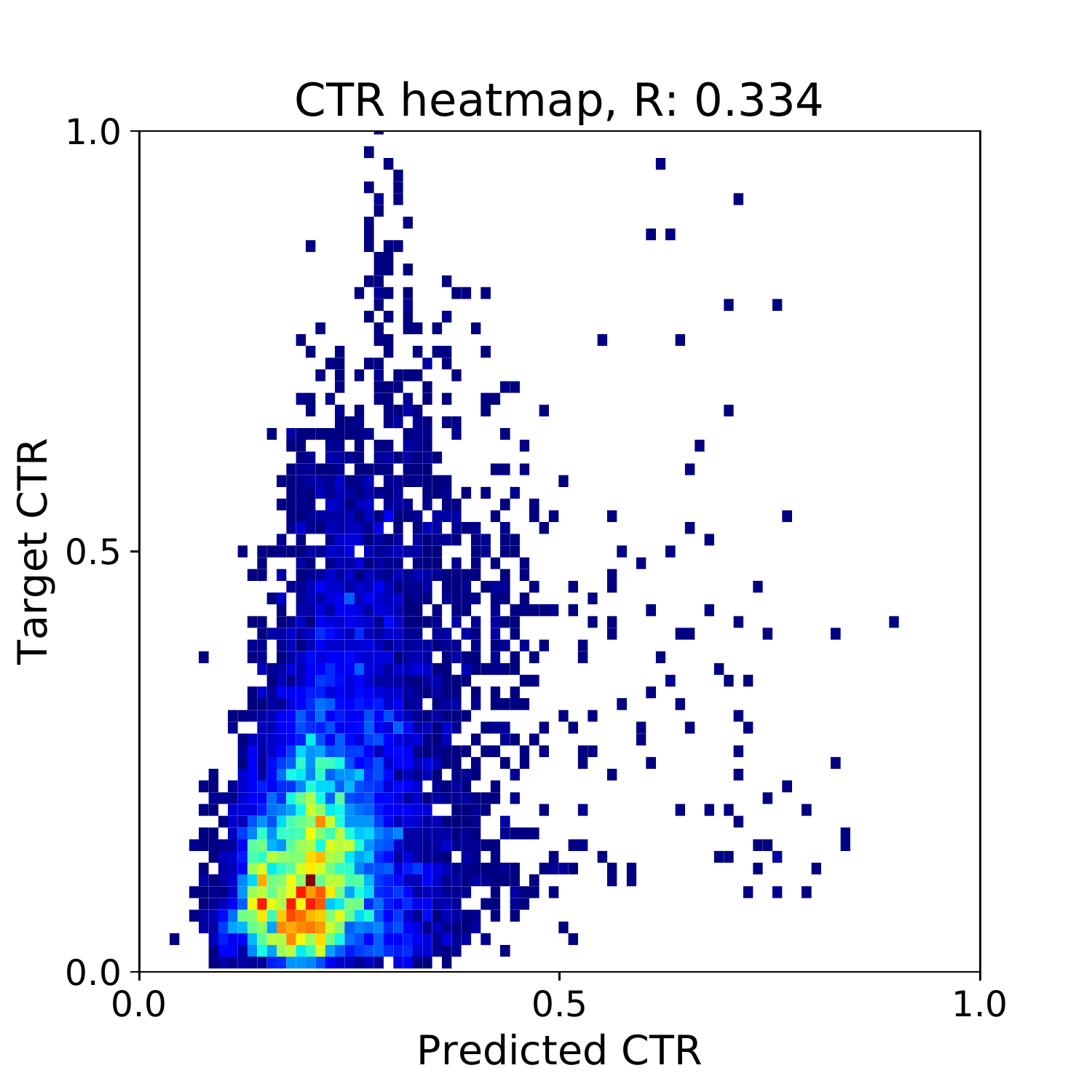}
      \begin{center}
        (c) Text
      \end{center}
      % \subcaption{Text}
      \label{fig:text}
    \end{minipage}

    % prediction
    \begin{minipage}{0.45\hsize}
      \centering
      \includegraphics[scale=0.39]{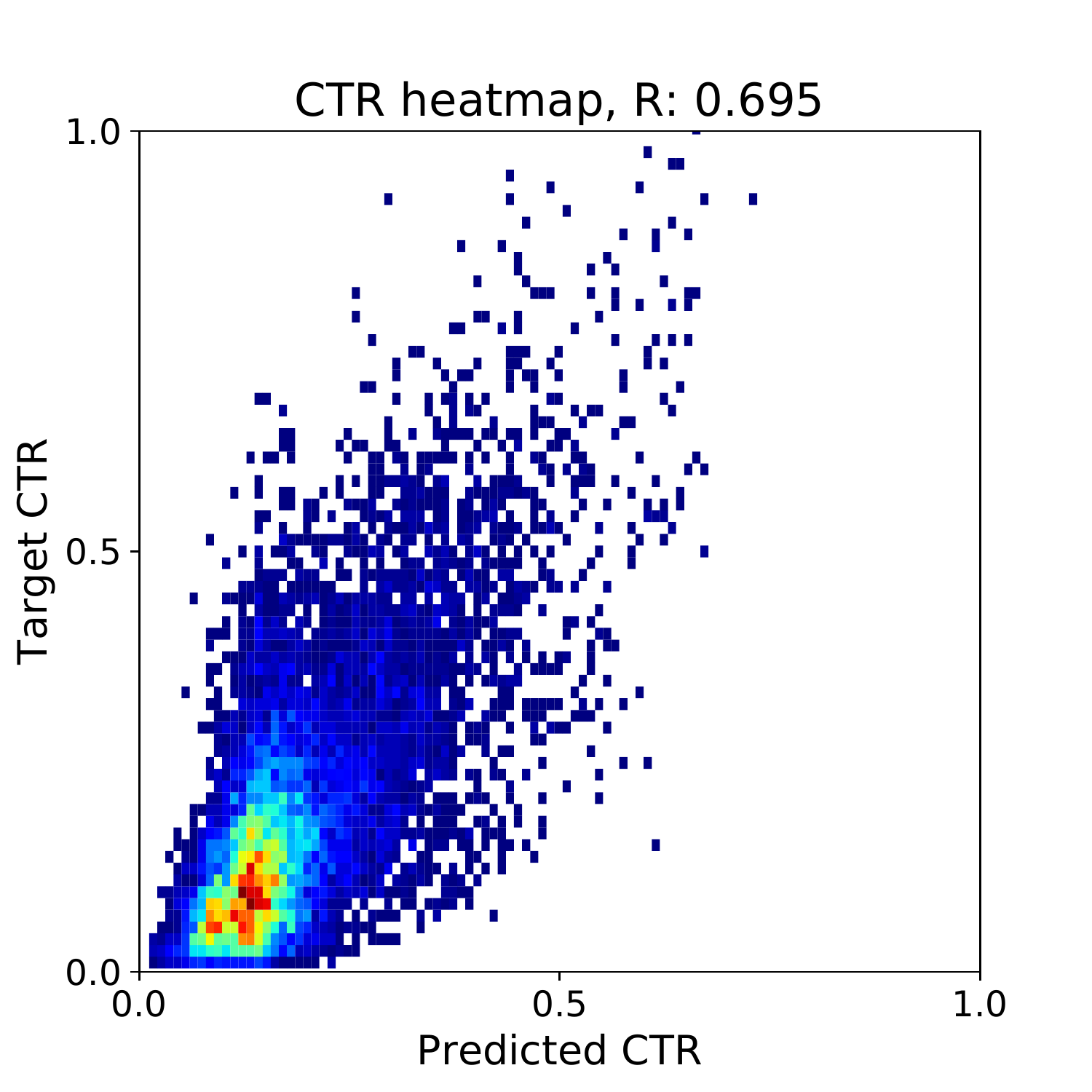}
      \begin{center}
        (d) Ours
      \end{center}
      % \subcaption{Ours}
      \label{fig:model}
    \end{minipage}
  \end{tabular}
  \caption{Scatter of the ground truth and predicted values of single feature models and our model.}
  % Upper left: Visual, upper right: Metadata, lower left: Text, lower right: Ours.
  \label{fig:scatter}
\end{figure*}

\section{Results}

In this section, we first compare our model with the baseline.
Then, we give the results of detailed ablation studies.
Finally, we show the visualized attention weights for better understanding of the proposed method.

The prediction accuracies of proposed methods and various comparison models are listed in Table~\ref{tab:results}.
We evaluate the prediction accuracies using two metrics, the root mean square error (RMSE) and the correlation coefficient (R).

\subsection{Comparison with baseline}

As the baseline, we use an modified model of Nakamura\cite{nakamura2018cnn} which includes visual and metadata feature extraction network (Fig.~\ref{fig:nakamura_model}), because there was no baseline of the CTR prediction model for online video advertisements.
We compared the baseline (ID:~13) with also the the modified proposed method (ID:~14) in which the text features are excluded from the input, because the baseline method has only visual and metadata input without text input.
The results demonstrated that the proposed method (ID:~14,15) had better prediction accuracy than the baseline.

\subsection{Ablation Studies}
\subsection*{Number of the sampling frames}
The number $n$ of the sampling frames is influential in the prediction accuracy.
When the number $n$ is larger, the visual information gets richer.
However, when the $n$ is too large, it might happen that many frames have similar information that could be be too much for the model.
ID:~11, 12, and 15 in Table~\ref{tab:results} show the performance improvement of the model from $n=10$ to $n=15$ (Ours), while deteriorating from $n=15$ to $n=20$.
Considering most of the video length is 15 seconds, one frame per second may be enough.
Besides, because most of the video length is 15 seconds, we consider it is balanced to have one frame per second.

\subsection*{Effectiveness of each modal input}
In order to evaluate the effectiveness of each modal input, we train the models ignoring each modal feature. 
The prediction accuracies obtained by the models is shown in Table~\ref{tab:results} (ID:~4-6).
Those results show each modal input contributes to the prediction.
Furthermore, ID:~1, 2, 3 in Table~\ref{tab:results} show the prediction results from the models which only consider a single modal feature.
The accuracy of metadata-based prediction model is the highest, while the accuracy of the text-based prediction model is the lowest.
Text data, e.g., the title and description of the advertisements, might have weaker effect on users' decision making on whether to click the advertisements or not than other modal data.
Fig.~\ref{fig:scatter} shows scatters of the prediction results, with the horizontal axis showing the predicted values and the vertical axis showing the ground truth.
For the text-based prediction, the predicted values tended to be fixed in a low range.
However, that was improved for the prediction using all data.

\subsection*{Effectiveness of separating input of metadata}
We separated the quantitative and qualitative variables so that they were normalized through the batch normalization layers.
ID:~7, 8, 9, and 15 in Table~\ref{tab:results} show we can achieve higher prediction accuracy by separating the metadata in the proposed way than by normalizing the quantitative metadata in advance or by both normalizing in advance and separating the metadata.
It is noteworthy that the proposed separation process exceeds the normalizing process, which is often used as a pre-procession of numerical data.

\subsection*{Effectiveness of additional layers for suppressing overfitting}
ID:~10 and 15 in Table~\ref{tab:results} show the additional layers for suppressing overfitting have certain effects to obtain the model which has better generalizability.

\begin{table}[t]
  \centering
  \caption{Correlation between each metadata and the CTR}
  \label{tab:correlation_meta_ctr}
  \begin{tabular}{lcc}
    \hline
    Key & correlation ratio & correlation coefficient \\ \hline
    promotion\_id & 0.443 & - \\
    publisher\_platform & 0.014 & - \\
    platform & 0 & - \\
    genre & 0.054 & - \\
    sub\_genre & 0.136 & - \\
    web\_app & 0.023 & - \\
    funnel & 0 & - \\
    creative\_type & 0 & - \\
    targeting\_type & 0.032 & - \\
    targeting\_gender & 0.002 & - \\
    targeting\_os & 0.030 & - \\
    targeting\_device & 0.030 & - \\
    targeting\_age\_min & - & 0.155 \\
    targeting\_age\_max & - & 0.110 \\
    target\_cost & - & -0.068 \\
    target\_cpa & - & -0.035 \\ \hline
  \end{tabular}
\end{table}

\begin{table}[t]
  \centering
  \caption{The prediction accuracies ignoring specific metadata input}
  \label{tab:withouteachmeta}
  \begin{tabular}{l|cc} \hline
   & \multicolumn{2}{c}{Metrics} \\
  Metadata input & RMSE & R \\ \hline
  Exc promotion\_id & 0.116 & 0.632 \\
  Exc publisher\_platform & 0.112 & 0.663 \\
  Exc platform & 0.110 & 0.674 \\
  Exc genre & 0.110 & 0.676 \\
  Exc sub\_genre & 0.110 & 0.676 \\
  Exc web\_app & 0.110 & 0.678 \\
  Exc funnel & 0.110 & 0.676 \\
  Exc creative\_type & 0.111 & 0.672 \\
  Exc targeting\_gender & 0.111 & 0.671 \\
  Exc targeting\_os & 0.115 & 0.642 \\
  Exc targeting\_device & 0.111 & 0.673 \\
  Exc targeting\_age\_min & 0.111 & 0.667 \\
  Exc targeting\_age\_max & 0.111 & 0.667 \\
  Exc target\_cost & 0.111 & 0.673 \\
  Exc target\_cpa & 0.112 & 0.661 \\ \hline
  All & \textbf{0.107} & \textbf{0.695} \\ \hline
  \end{tabular}
\end{table}

\begin{table}[t]
  \centering
  \caption{The prediction accuracies ignoring specific text input}
  \label{tab:withouteachtext}
  \begin{tabular}{l|cc}
  \hline
   & \multicolumn{2}{c}{Metrics} \\
  Text input & RMSE & R \\ \hline
  Exc advertiser\_name & 0.108 & 0.695 \\
  Exc account\_name & \textbf{0.107} & \textbf{0.698} \\
  Exc promotion\_name & 0.109 & 0.681 \\
  Exc creative\_title & 0.108 & 0.688 \\
  Exc creative\_description & \textbf{0.107} & \textbf{0.697} \\ \hline
  All & 0.107 & 0.695 \\ \hline
  \end{tabular}
\end{table}

\begin{table*}[!h]
  \centering
  \caption{Average attention weights of each frame feature.}
  \label{tab:frame_attention_weight}
  \begin{tabular}{lccccccccccccccc}
    \hline
    Frame No. & \textbf{1} & \textbf{2} & \textbf{3} & 4 & 5 & 6 & \textbf{7} & 8 & 9 & 10 & 11 & 12 & 13 & 14 & \textbf{15} \\ \hline
    Avg. weight & \textbf{0.103} & \textbf{0.087} & \textbf{0.085} & 0.061 & 0.047 & 0.055 & \textbf{0.076} & 0.051 & 0.057 & 0.067 & 0.058 & 0.070 & 0.060 & 0.048 & \textbf{0.073} \\ \hline
  \end{tabular}
\end{table*}

\subsection{Correlations between input features and prediction accuracy}
To analyze the effect of detailed features, we measured the correlation between each metadata and the CTR, i.e., the correlation ratios for qualitative metadata, and the correlation coefficients for quantitative metadata (Table~\ref{tab:correlation_meta_ctr}).
We can find that {\it promotion\_id} has a relatively strong correlation with the CTR (0.443), and {\it sub\_genre}, {\it targeting\_age\_min}, and {\it targeting\_age\_max} have weak correlations (0.136, 0.155, and 0.110, respectively), and most features have zero or very low correlation.

Then, we explored if the prediction accuracy was improved by ignoring specific metadata or text input features.
First, we excepted each metadata from the input and predicted the CTR, separately.
The prediction accuracies are shown in Table~\ref{tab:withouteachmeta}. 
The results, e.g. {\it Exc platform}, {\it Exc funnel}, and {\it Exc creative\_type}, indicate ignoring a specific metadata feature doesn't improve the prediction accuracy even though the ignored metadata has zero correlation with the CTR.
The results also indicate that the {\it promotion\_id} has the strongest contribution to the prediction accuracy because when {\it promotion\_id} is ignored, the accuracy is the worst.
Second, we excepted each text feature from the input and predicted the CTR.
The prediction accuracies are shown in Table~\ref{tab:withouteachtext}.
We found the prediction accuracy was improved by ignoring {\it account\_name} or {\it creative\_description}, and we confirmed the accuracy was much improved when both features were ignored.
In summary, we can improve the prediction accuracy by ignoring specific features. 
However, we cannot specify which features should be ignored in advance. 
Therefore, to optimize the model to a specific dataset, we have to explore which features should be ignored.

\subsection{Visualizing attention weights}
\begin{figure}[!t]
  \centering
  \includegraphics[scale=0.38]{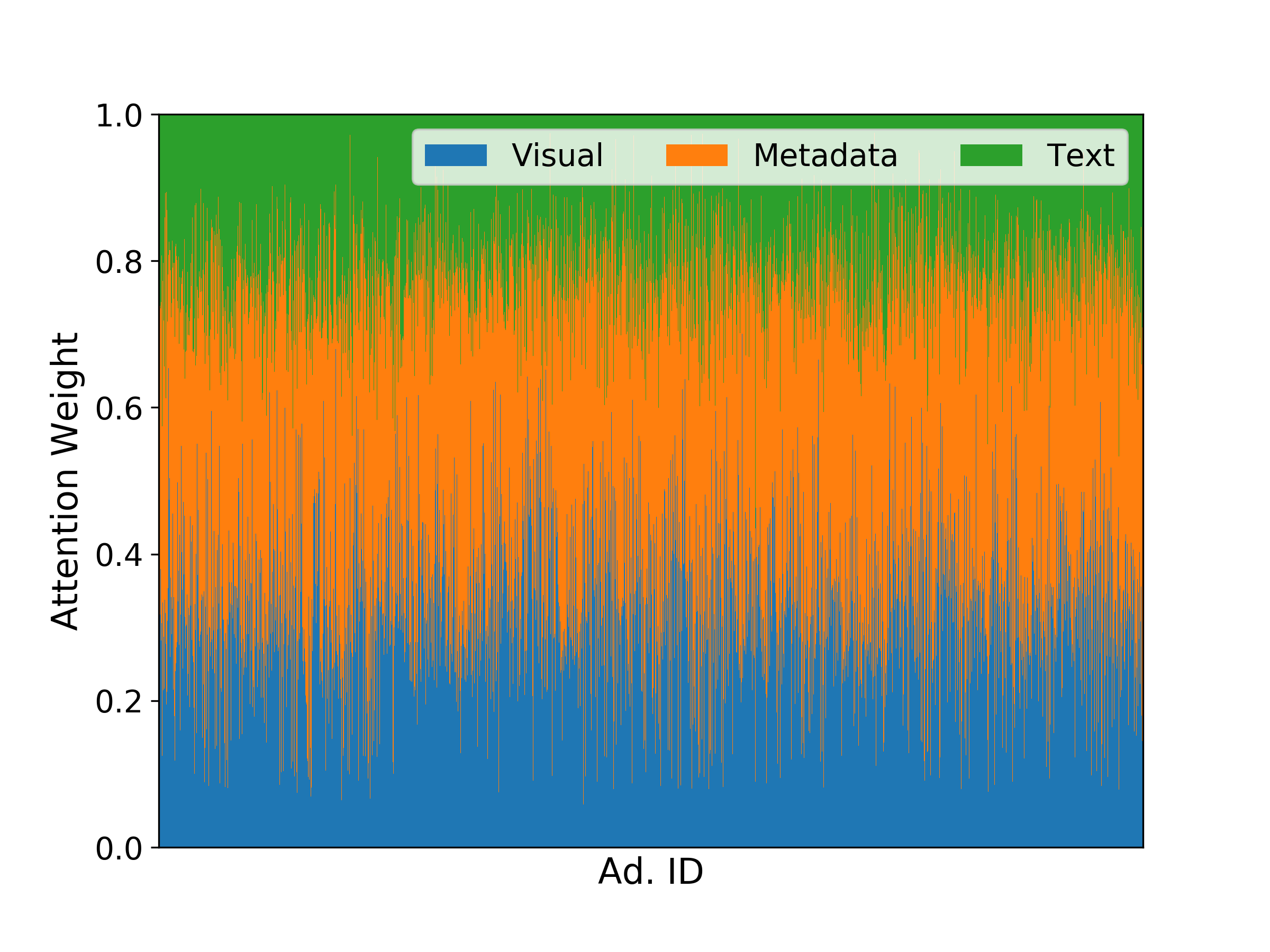}
  \caption{Attention weights of each modal feature.}
  \label{fig:modal_attention}
\end{figure}

We visualized the attention weight of each modal feature of all data in the test dataset.
In Fig.~\ref{fig:modal_attention}, the horizontal axis is the advertisement ID and the vertical axis is the stacked graph representing the weight of attention.
Blue bars represent the visual, orange bars represent the metadata, and green bars represent the attention weights of the text.
In order to compare the weights of each modal feature quantitatively, the average weights of each modal feature were calculated and are shown in Table~\ref{tab:modal_attention_weight}.
The average weight of metadata was the highest, while that of the text was the lowest. This implies that on average, the model places the most attention on metadata, and it is the primary contributor to the CTR.

\begin{figure}[!t]
  \centering
  \includegraphics[scale=0.38]{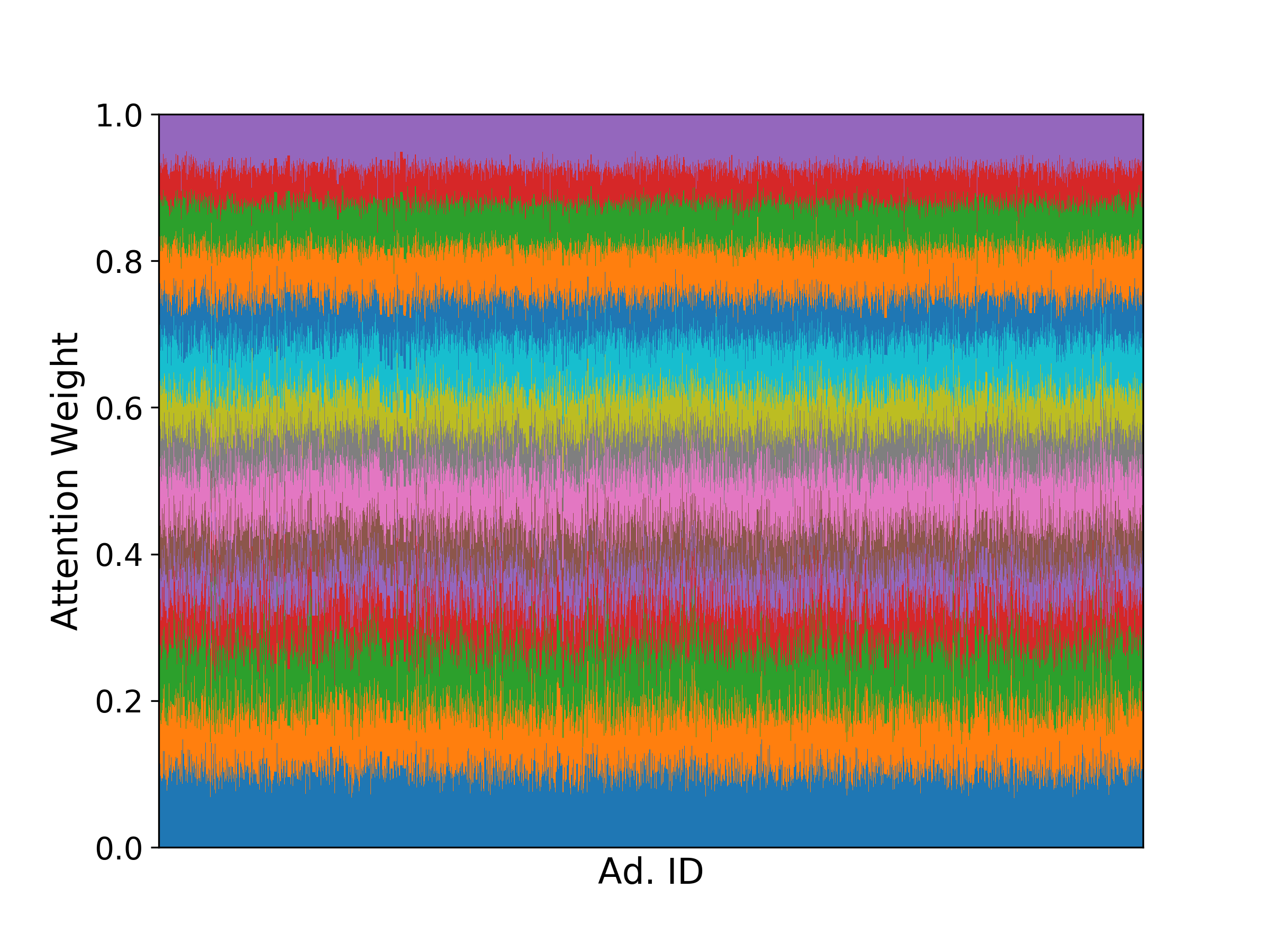}
  \caption{Attention weights of each frame feature (15 frames of each video).}
  \label{fig:frame_attention}
\end{figure}

Next, we visualized the attention weights of each frame feature of all data in the test dataset.
In Fig.~\ref{fig:frame_attention}, the horizontal axis is the advertisement ID and the vertical axis is the stacked graph representing the attention weights in the frame order, such that the weight of the first frame is at the bottom and that of the last frame is at the top.
The average weights of each frame feature were calculated and are shown in Table~\ref{tab:frame_attention_weight}.
The average weight of the first frame is the highest, while those of the second, third, seventh, and the final frames are relatively higher than others.
This implies that on average, the model places attention on the first few seconds and the last scene of the video, and these frames primarily contribute to the CTR.

\begin{table}[!t]
  \centering
  \caption{Average attention weights of each modal feature.}
  \label{tab:modal_attention_weight}
  \begin{tabular}{lccc}
    \hline
    Modal & Visual & Metadata & Text \\ \hline
    Avg. Weight & 0.321 & 0.452 & 0.226 \\
    \hline
  \end{tabular}
\end{table}

\section{Conclusions}
In this research, we proposed a method for predicting the CTR of online video ads and analyzing the factors affecting the CTR.
Based on an existing method that predicts impressions of TV commercials, we optimized the network and parameters to online video ads.
We separated the qualitative variables and quantitative variables to input into the FC and BN layers, respectively, and added normalization layers to the networks in order to suppress overfitting.
As a result, in terms of the prediction accuracy, a CORR of 0.695 was obtained.
Furthermore, by visualizing the attention weights, it was demonstrated that metadata largely affect the CTR, and the first few seconds and last scene of a video relatively contributed to the determination of the CTR.

\IEEEtriggeratref{13}


\bibliographystyle{IEEEtran}
\bibliography{report}


% \end{thebibliography}

\end{document}